\documentclass[12pt,a4paper]{article}
\usepackage[utf8]{inputenc}
\DeclareUnicodeCharacter{202F}{ }
\usepackage[T1]{fontenc}
\usepackage{graphicx}           
\graphicspath{{/Users/pengyusen/Downloads/latexpic/}}
\usepackage[authoryear]{natbib}
\usepackage{hyperref}
\usepackage{float}
\usepackage{amsmath}
\usepackage{amsmath,amssymb}
\usepackage{url}
\usepackage{booktabs}
\usepackage{tabularx}
\usepackage{caption}

\usepackage{titlesec}

\titleformat{\paragraph}[block]
{\normalfont\normalsize\bfseries}  
{}                                 
{0pt}                              
{}                                 

\usepackage{etoolbox}

\newenvironment{keywords}{
	\vspace{1em}%
	\noindent\textbf{Keywords:}\quad%
}{\par}

\title{E.A.R.T.H.: Structuring Creative Evolution through Model Error in Generative AI}

\author{%
	\centering
	\begin{minipage}{0.8\textwidth}
		\centering
		\small
		Yusen Peng\textsuperscript{1}, Shuhua Mao\textsuperscript{2}\\[0.5ex]
		\textsuperscript{1}University of Warwick\\
		\textsuperscript{2}Wuhan University of Technology\\[0.5ex]
		\texttt{Yusen.Peng@warwick.ac.uk}, \texttt{maosh\_415@whut.edu.cn}
	\end{minipage}
}

\date{}   

\begin{document}
	\maketitle

\label{key}	
\begin{abstract}
		\noindent How can AI move beyond imitation toward genuine creativity? This paper proposes the E.A.R.T.H.\ framework, a five-stage generative pipeline that transforms model-generated errors into creative assets through Error generation, Amplification, Refine selection, Transform, and Harness feedback. Drawing on cognitive science and generative modeling, we posit that “creative potential hides in failure” and operationalize this via structured prompts, semantic scoring, and human-in-the-loop evaluation. Implemented using LLaMA-2-7B-Chat, SBERT, BERTScore, CLIP, BLIP-2, and Stable Diffusion, the pipeline employs a composite reward function based on novelty, surprise, and relevance. At the Refine stage, creativity scores increase by 52.5\% (1.179 to 1.898, $t=-5.56$, $p<0.001$), with final outputs reaching 2.010—a 70.4\% improvement. Refined slogans are 48.4\% shorter, 40.7\% more novel, with only a 4.0\% drop in relevance. Cross-modal tests show strong slogan-to-image alignment (CLIPScore: 0.249; BERTScore F1: 0.816). In human evaluations, the generated outputs were consistently rated highly, demonstrating strong creative quality and expressive clarity. Feedback highlights stylistic precision and emotional resonance. These results demonstrate that error-centered, feedback-driven generation enhances creativity, offering a scalable path toward self-evolving, human-aligned creative AI.
\end{abstract}


\section{Introduction: When AI Errors Become Creative Opportunities}

Generative AI systems such as GPT‑4, DALL·E, and Stable Diffusion are rapidly transforming creative practices across diverse domains—from advertising and branding to scientific ideation and conceptual design. These models produce text, images, music, and novel concepts that increasingly blur the boundary between human and machine imagination. However, they are frequently criticized for generating hallucinations, logical inconsistencies, or low‑confidence content, commonly labeled as \emph{errors} and subsequently suppressed through supervised alignment or reinforcement learning from human feedback \citep[e.g.,][]{Ouyang2022}.

This paper adopts a radically different perspective: these errors are not mere artifacts of failure, but underutilized sources of machine creativity.

Historically, human creativity has often emerged not from perfect execution but from deviation, surprise, or the unexpected. Alexander Fleming’s accidental discovery of penicillin (1929), the surrealists’ fascination with perceptual misrecognition, and the dissonant improvisations of jazz all exemplify how disruption and deviation can catalyze breakthrough ideas. Contemporary studies in AI reinforce this view: \citet{Haase2023} show that highly surprising outputs from language models are frequently rated as more creative, and \citet{DoshiHauser2024} demonstrate that flawed or low‑confidence responses can improve the novelty–utility tradeoff in ideation tasks. In domains such as drug discovery, hallucinated molecular structures have even yielded viable new compounds \citep{MarkTechPost2025}.

Yet the central challenge remains: how can this error‑driven potential be systematically activated, rather than incidentally stumbled upon?

To address this, we propose the E.A.R.T.H.\ framework, a five‑stage generative architecture encompassing error generation, amplification, refine selection, transform, and harness feedback. Rather than treating model errors as undesirable noise, the framework treats them as raw creative material—extracting, amplifying, recontextualizing, and evolving them across structured stages. Drawing inspiration from predictive coding \citep{Friston2009}, compression‑driven intrinsic motivation \citep{Schmidhuber2009}, and surprise‑based search dynamics \citep{Yannakakis2016}, we argue that errors are not obstacles but friction points that can spark generative imagination.

Through an empirical pipeline involving over 500 slogan generations, we demonstrate that low‑likelihood or semantically divergent outputs—when systematically filtered, ranked, and reconstructed—can produce more novel, metaphor‑rich, and emotionally resonant content than conventional high confidence outputs. For example, a discarded fragment such as “the night melts in his palm,” initially filtered out by conventional scoring, can become the conceptual seed of a compelling technology advertisement when reframed through contextual prompts and cross‑modal grounding.

In short, AI errors can function as conceptual catalysts: not dead ends, but starting points for creative evolution. Just as penicillin emerged from an accidental contaminant, the next generation of machine‑generated creativity may arise not in spite of mistakes but because of them.

	\section{Re‑examining Creativity and Errors: Theoretical and Technical Foundations}
	
	To determine whether AI‑generated “errors” can genuinely be regarded as creative, we must begin with a foundational question: What constitutes creativity? Across cognitive science and computational creativity research, creativity is commonly defined by three essential criteria: novelty, unexpectedness, and value \citep{Kern2024,Abraham2025}. Creative outputs must be original, deviate from expectation, and generate aesthetic, emotional, or functional resonance. This triadic standard underpins not only human creativity but also forms the evaluative backbone of our proposed E.A.R.T.H.\ framework.
	
	In this section, we first revisit how these core criteria apply in the context of generative AI, then examine the internal mechanisms that give rise to errors, and finally synthesize a theoretical foundation for error‑driven creativity that guides our system design.
	
	\subsection{The Nature of Creativity and Its Applicability in AI}
	
	Margaret Boden \citep{Boden2024} distinguishes three categories of creativity: combinational, exploratory, and transformational. Generative models like LLaMA‑2 and GPT‑4 already excel at combinational creativity—reconfiguring existing elements in novel ways. However, under certain sampling strategies or during semantic drift, these systems can incidentally produce outputs that cross into exploratory or even transformational space: generating content that challenges conventions, alters conceptual boundaries, or reframes meaning.
	
	Abraham \citep{Abraham2025} warns against reducing creativity to surface level novelty and emphasizes the role of subjective experience—including affective impact, interpretive surprise, and aesthetic insight. While current AI lacks intrinsic subjectivity, its outputs can nonetheless provoke human subjective responses. In our framework, these responses serve as external validators—particularly in Stage H, where human feedback provides the final measure of creative value.
	
	Interestingly, many generative errors—such as semantic incongruity, rhetorical rupture, or unexpected metaphor—align closely with the very criteria used to assess creativity. If properly captured and reframed, these anomalies offer structural affordances for conceptual leaps. Understanding how such deviations arise is thus crucial for operationalizing error as creative input, which the E.A.R.T.H.\ process systematically attempts to achieve.
	
	\subsection{The Origin of Generative AI Errors}
	
	Errors in generative AI stem primarily from the stochastic, probabilistic nature of large language models (LLMs). Unlike deterministic systems, autoregressive models like LLaMA‑2 generate sequences by sampling from probability distributions over tokens. Techniques such as temperature scaling, top‑k, and nucleus (top‑p) sampling \citep{Holtzman2020} are designed to enhance diversity, but in doing so they inevitably introduce low‑likelihood outputs—many of which appear as “mistakes,” yet may contain high creative potential.
	
	Compounding this, the training corpus itself contains stylistic conflicts, incomplete logic, and semantic ambiguities. These latent tensions re‑emerge during generation, manifesting as hallucinations, conceptual blends, or stylistic mismatches. Rather than being dismissed outright, our approach treats these outputs as potential seeds for creative recomposition—especially in the Amplify and Refine stages, where semantic surprise and divergence are scored and preserved.
	
	Additionally, the architecture of LLMs encourages cascade effects: a single unpredictable token may lead the model down a novel trajectory, generating content that—while deviant—resembles the associative jumps found in poetry or surrealist art. While RLHF alignment procedures reduce harmful outputs, they also constrain stylistic risk‑taking. \citet{Ouyang2022} show that such preference‑based fine‑tuning often flattens the expressive space, limiting the system’s capacity for generative innovation.
	
	In short, AI “errors” are not bugs but statistical excursions—probabilistic deviations that, under the right framing, expose untapped creative directions.
	
	\subsection{A Theoretical Framework for Error‑Driven Creativity}
	
	To elevate such deviations from noise to creative resource, we draw upon interdisciplinary cognitive‑computational theories that regard prediction violations as sites of learning and novelty:
	
	\begin{itemize}
		\item \textbf{Predictive Coding} \citep{Friston2009}: Conceptualizes the brain as a prediction machine. Deviations from expectation produce surprisal, prompting model updates. Analogously, low‑likelihood outputs in AI signal model uncertainty and point to zones of conceptual novelty.
		\item \textbf{Compression Progress Theory} \citep{Schmidhuber2009}: Rewards learning systems for discovering order within randomness. Errors that initially appear incoherent but later reveal structure are particularly valuable—echoing how our Refine and Transform stages reward syntactically deviant but semantically meaningful outputs.
		\item \textbf{Surprise Search} \citep{Yannakakis2016}: Favors trajectory deviations—outcomes that diverge not merely from the norm but from immediate expectation. This model underpins our scoring in the Amplify and Refine stages, where surprise is treated as signal, not defect.
	\end{itemize}
	
	These frameworks collectively affirm that novelty alone is insufficient. What matters is surprise grounded in internal expectations—precisely the signal encoded in our scoring system via semantic distance, log‑likelihood, and relevance constraints.
	
	\subsection{Toward an Integrated View: Error as Structured Creative Fuel}
	
	Together, these theories support a unified claim: errors are structured and learnable, not random noise. Whether interpreted as prediction violations, compressive outliers, or semantic drifts, they reveal hidden creative potential that becomes actionable through a staged refinement system.
	
	In our experimental pipeline, such errors are amplified, reranked, and recontextualized—culminating in outputs that outperform conventional generations in human‑rated creativity, emotional resonance, and symbolic expressiveness. Rather than filtering out anomaly, we design the system to retain, reshape, and evolve it—establishing a reproducible link between deviation and invention.
	
	By bridging theoretical insight with computational design, the E.A.R.T.H.\ framework demonstrates how errors can be recast as structured creative fuel, offering a path toward human–AI co‑creation grounded not in precision but in productive imperfection.

\section{The E.A.R.T.H.\ Framework—Operationalizing Error-Driven Creativity in Generative AI}

Generative AI systems have long been committed to optimizing accuracy and controllability, and avoiding “illusions” and low-confidence content as much as possible. However, this study proposes a subversive perspective: errors are not impurities that the system should exclude, but may become the source of creativity. Based on cognitive theories such as predictive coding, compression-driven intrinsic motivation (Compression Curiosity), and surprise-based evolutionary exploration, and drawing on the prompt –response–reward structure proposed by \cite{Huang and Rust2024}, this study constructed a modular five-stage creative system: the E.A.R.T.H.\ framework. This framework forms a systematic “creation from errors” mechanism by guiding, identifying, screening, converting and harnessing feedback on error outputs.

\begin{figure}[ht]
	\centering
     \includegraphics[width=1.0\linewidth]{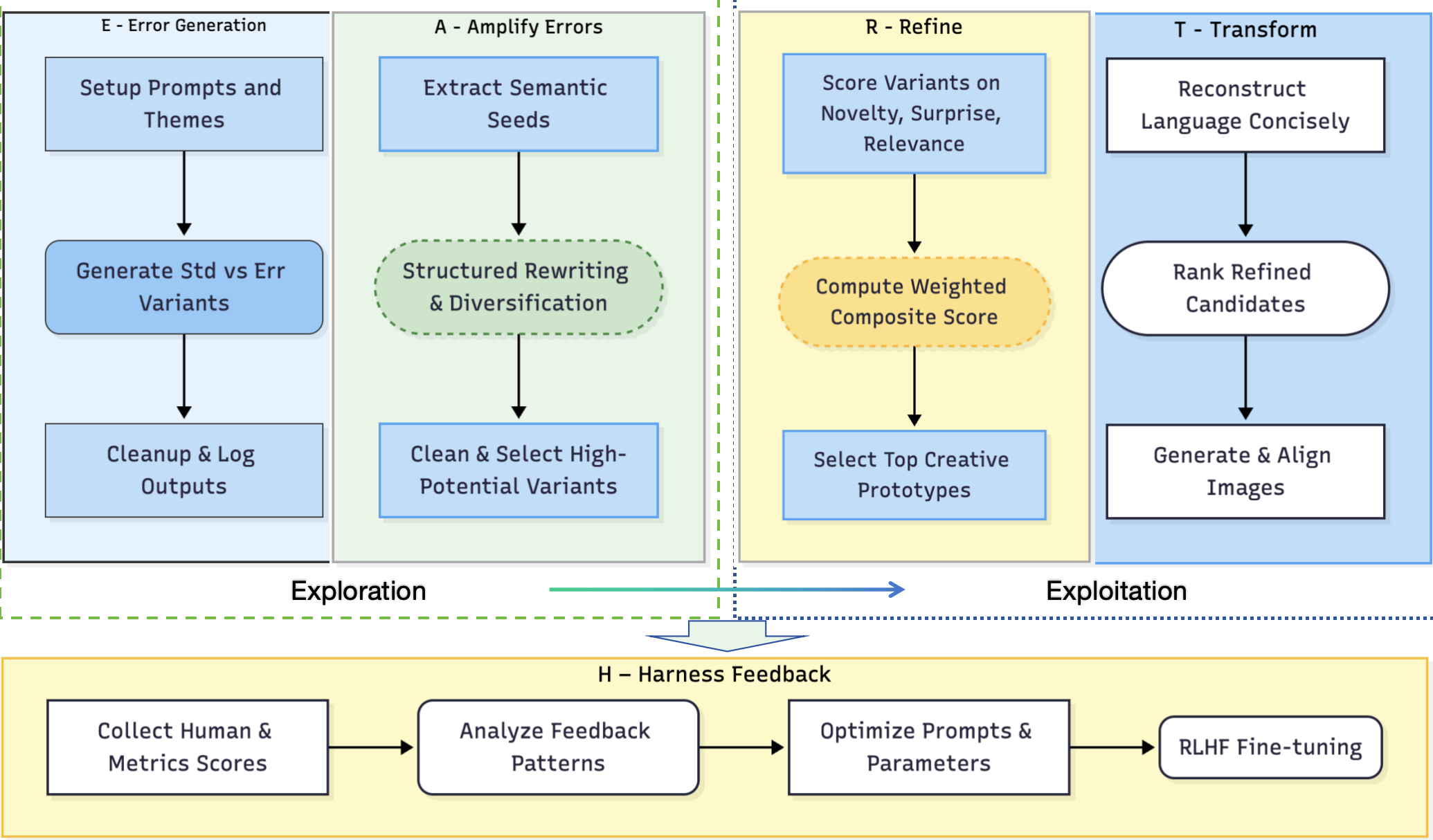}
		\caption{The E.A.R.T.H. framework: A feedback-driven model for error-based generative creativity. }
	\label{fig:earth-framework}
\end{figure}
	
\subsection{E – Error Generation: Intentionally Introducing Nonlinear Deviations via Sampling Strategies}

\subsubsection{Objectives and Motivation}

In traditional text generation paradigms, language models are typically trained and deployed under strategies that maximize the avoidance of errors and suppression of low‑confidence outputs. The decoding process often adopts low‑temperature sampling or greedy decoding to produce fluent, controllable, and semantically stable texts. However, while this ``bias‑removal'' strategy improves acceptability, it also severely suppresses the creative potential of language models—particularly their ability to explore the low‑probability ``long tail'' of the output distribution.

In the E.A.R.T.H.\ framework, we deliberately subvert this conventional path by treating ``errors'' as fuel for creative generation rather than noise. We explicitly guide the model to deviate from the mainstream output distribution through decoding‑level interventions to discover potential unexpectedness and innovation. As Friston noted, ``Errors are outliers in model learning,'' and it is precisely these outliers that may lead systems toward new cognitive pathways \cite{Friston2009}.

\subsubsection{Methodological Support and Technical Principles}

Our induction of ``errors'' does not rely on modifying training data or intervening in model architecture, but is entirely performed through decoding‑level parameter control. This approach is supported by theoretical and empirical validations from multiple top‑tier conference papers:

\begin{enumerate}
	\item Nucleus Sampling (Top‑p Sampling)\\
	Holtzman et al.\ (\cite{Holtzman2020}) proposed nucleus sampling, which selects from the smallest set of tokens whose cumulative probability exceeds $p$. This effectively mitigates the ``text degeneration'' problem and extends the boundary of diversity while maintaining generation quality \cite{Holtzman2020}.
	
	\item Temperature Scaling\\
	Ippolito et al.\ (\cite{Ippolito2019acl}) demonstrated in a comparative study that increasing the softmax temperature $\tau$ significantly enhances the randomness of model sampling, making the output more unpredictable and creatively divergent, especially when combined with nucleus sampling \cite{Ippolito2019acl}.
	
	\item Diversity‑Promoting Objectives\\
	Li et al.\ (\cite{Li2016}) introduced the Maximum Mutual Information (MMI) objective, emphasizing the maximization of information complementarity between inputs and outputs during training, thereby improving semantic richness and variation. This provides a theoretical foundation for our decoding‑level ``non‑optimal output guidance'' \cite{Li2016}.
\end{enumerate}

Collectively, these methods demonstrate that increasing temperature and combining with top‑p truncation can significantly enhance the divergence and creative potential of generated content while retaining linguistic quality.

\subsubsection{Sampling Implementation and Parameter Settings}

Based on the above methodological principles, we designed a temperature ‑controlled parameter branching strategy in the actual generation process to systematically control the balance between stability and creativity of the outputs. Specifically, we implemented two sampling approaches:

\begin{itemize}
	\item Standard Sampling (\texttt{std}): with temperature $\tau = 0.7$ and top‑p = 0.9, to generate semantically stable and structurally normative advertising slogans.
	\item Error‑Induced Sampling (\texttt{err}): with temperature $\tau = 1.3$ and top‑p = 0.9, to significantly enhance diversity and unpredictability.
\end{itemize}

In addition, we retained two placeholder baseline methods (CAN and DQD) to facilitate subsequent comparative analysis across different generation strategies.

As shown in Figure~\ref{fig:earth-framework} Stage E , the overall procedure includes five steps:

\begin{enumerate}
	\item System Prompt Setup: define the model as a “creative advertising copywriter” to guide it toward generating short, memorable slogans.
	\item Thematic Prompt Construction: structure input queries around five fixed themes (e.g., “Self‑Transcendence,” “Green Future”).
	\item Parameter Branching and Diverse Generation: for each theme, perform both standard and error‑induced sampling, generating five slogans per method.
	\item Auto‑Cleaning and Trimming: remove redundant prefixes and retain only the main slogan text.
	\item Structured Logging: store all sample information, including generation method (\texttt{method}), input prompt, and output slogan. The results are recorded in a standardized CSV format for downstream analysis and evaluation.
\end{enumerate}

\subsubsection{Qualitative Results Analysis}

A total of 100 slogans were generated in this stage, with each method (std, err, CAN, DQD) contributing 25 slogans. After trimming, all texts were made comparable.

As shown in Table~\ref{tab:qualitative}, the Error‑Induced Sampling method (\texttt{err}) demonstrates a significantly higher level of creative leap compared to standard sampling (\texttt{std}):

\begin{itemize}
	\item Use of more metaphorically bold expressions (e.g., ``Ignite Your Soul'');
	\item Stronger visual imagery (e.g., ``reach for the stars'');
	\item Broader conceptual combinations (e.g., blending ``ritual'' with ``technology'').
\end{itemize}

In contrast, standard sampling tends to produce templated sentence structures, frequently repeating generic phrases like ``Unlock Your True\dots'', limiting creative space.

\captionsetup[table]{justification=raggedright,singlelinecheck=false}

\begin{table}[ht]
	\centering
	\caption{Comparative examples of slogans generated via standard and error-induced sampling.}
	\label{tab:qualitative}
	\begin{tabularx}{\textwidth}{@{}l X@{}}
		\toprule
		Method & Slogan Examples \\
		\midrule
		\texttt{std} &
		\begin{minipage}[t]{\hsize}
			Unlock Your True Potential: Transcend Limits, Achieve Greatness!\\
			Rise Above Your Limits, Transcend Your Self.\\
			Unleash Your True Self: Go Beyond the Ordinary!
		\end{minipage} \\
		\addlinespace
		\texttt{err} &
		\begin{minipage}[t]{\hsize}
			Go beyond the ordinary, reach for the stars.\\
			Elevate Beyond the Ordinary.\\
			Elevate Your Life. Ignite Your Soul.
		\end{minipage} \\
		\bottomrule
	\end{tabularx}
\end{table}

The above results validate the core hypothesis of this stage: by simply adjusting sampling hyperparameters, it is possible to induce the language model to generate text outputs with significantly different styles, greater structural leaps, and higher creative potential. These non‑mainstream candidates obtained through “error‑inducing” strategies provide a rich foundation for exploration in subsequent stages such as A (Amplify) and R (Refine), serving as the critical first step in realizing a creativity mechanism.

\subsection{A – Amplify Errors: Expanding Creative Deviations via Controlled Regeneration}

The Error‑Induction Stage (E) introduces creative disruption through high ‑temperature sampling, producing outputs that deviate from normative syntax and semantics. However, not all such deviations are inherently valuable. The Amplification Stage (A) takes a more selective approach: instead of accepting all anomalies, it extracts and expands only those that exhibit internal coherence, metaphorical potential, or structural novelty. In this stage, we aim to turn incidental divergence into structured creativity by building upon promising “semantic seeds.”

Although all three core stages—Amplify (A), Refine (R), and Transform (T)—involve creativity scoring, the scoring formulas differ in both purpose and construction, and such differences influence downstream outputs in form, style, and structure. To ensure interpretive clarity and methodological transparency, we first delineate and contrast the three scoring functions used across stages:

\subsubsection{Stage A (Amplify) -- Creativity Score (Seed Selection)}

Used to select top-\(k\) semantic seeds from error‑induced outputs, this stage adopts a four‑component scoring metric:

\begin{equation}
\mathit{CreativityScore}
= 1.0\cdot\mathit{Novelty}
+ 0.5\cdot\mathit{Surprise}
+ 0.5\cdot\mathit{Divergence}
+ 0.2\cdot\mathit{Relevance}
  \label{eq:creativity-score}
\end{equation}

This weighting reflects the exploratory emphasis of the A‑stage, prioritizing semantic deviation (novelty) and stylistic irregularity (divergence) as indicators of latent creative value. It is not used for final selection, but rather to pre‑screen for amplification.

\subsubsection{Stage R (Refine) -- R Score (Primary Creativity Filter)}

At this stage, a simplified three‑dimensional score is employed to evaluate and select among the 75 amplified variants:

\begin{equation}
	R_{\mathrm{score}}
	= 0.4\,\mathit{Novelty}
	+ 0.4\,\mathit{Surprise}
	+ 0.2\,\mathit{Relevance}
	\label{eq:R-score}
\end{equation}

This formula balances novelty and surprise—aligned with classic creativity theories—while ensuring semantic coherence via a reduced relevance weight. The R‑score thus serves as the main selection criterion for creative prototype extraction.

\subsubsection{Stage T (Transform) -- Final Selection Score}

Unlike the R‑stage, which evaluates a broad pool, the T‑stage selects one final output among refined candidates. As stylistic variation has already been introduced, the final selection prioritizes semantic clarity and formal compactness:

\begin{equation}
	T_{\mathrm{score}}
	= 0.7\,\mathit{Novelty}
	+ 0.3\,\mathit{Relevance}
	\label{eq:T-score}
\end{equation}

Here, surprise is intentionally excluded: T‑stage outputs are deterministically rewritten and evaluated on expression rather than model uncertainty. The formula reflects the transition from creativity evaluation to communicative optimization, emphasizing expressive precision and metaphorical clarity.

\subsubsection{Seed Extraction: Identifying Expandable Creative Units}

The first sub‑stage in A involves scoring all outputs from the \texttt{err} generation method of Stage E to identify candidates worthy of creative amplification. We define a composite metric called \emph{CreativityScore}, which integrates four dimensions:

\begin{itemize}
	\item Novelty: measured as \(1 - \cos(\mathbf{e}_{\text{prompt}}, \mathbf{e}_{\text{output}})\) between SBERT embeddings of the prompt and generated text, reflecting semantic distance and the model’s ability to explore conceptually distant territory.
	\item Surprise: operationalized as the mean negative log‑likelihood from the LLaMA‑2‑7B‑Chat model, capturing the model’s uncertainty and deviation from internal expectations.
	\item Divergence: calculated using Jensen–Shannon divergence between the token distributions of the prompt and output, indicating shifts in syntactic or stylistic structure.
	\item Relevance: measured using BERTScore F1 between the generated slogan and its prompt, ensuring that outputs remain meaningfully grounded in the original theme.
\end{itemize}

We then compute:

\begin{equation}
	\mathit{CreativityScore}
	= 1.0\cdot\mathit{Novelty}
	+ 0.5\cdot\mathit{Surprise}
	+ 0.5\cdot\mathit{Divergence}
	+ 0.2\cdot\mathit{Relevance}
	\label{eq:creativity-score 1}
\end{equation}

Using this scoring, we select the top 15 outputs as “semantic seeds”—not final slogans, but expandable starting points with promising creative potential. This stage is not intended as a terminal evaluation but as a pre‑screening mechanism to support controlled creative expansion in the next phase.

\subsubsection{Creative Amplification: Structured Rewriting with Diversity Control}

After seed extraction, we move to controlled regeneration, where each semantic seed is rewritten to produce stylistically varied, punchier outputs. We construct structured prompts in the following dialogue format:

\begin{quote}
	System: You are a creative advertising copywriter. Produce exactly one concise slogan.\\
	User: <semantic seed>
\end{quote}

We use the \texttt{LLaMA-2-7B-Chat} model with the following parameters:
\begin{itemize}
	\item Temperature = 1.5
	\item Top-p = 0.95
	\item Max new tokens = 55
	\item Variants per seed = 5
\end{itemize}

This configuration enables expressive diversity while constraining semantic collapse. The regeneration process yields 75 amplified slogans (15 seeds~$\times$~5 variants), which are then cleaned to remove generic prefix artifacts (e.g., “Sure! Here’s your slogan…”), retaining only the core output for downstream analysis.

This controlled amplification design is inspired by recent advances in compositional generation chains and creativity modeling. It reflects three key practices:
\begin{itemize}
	\item \textbf{Semantic recombination}, akin to Compositional Fine‑Tuning \citep{Bursztyn2022};
	\item \textbf{Prompt chaining and rewriting}, drawing on finetuned language models as zero‑shot learners \citep{Wei2023};
	\item \textbf{Output reuse for human‑like revision}, as in explainable self‑referential revision frameworks \citep{Li2021}.
\end{itemize}

\subsubsection{Conceptual Significance and Design Implications}

The Amplify Errors stage plays a crucial transitional role in the creative evolution framework. It transforms raw, unstructured “errors” into diversified expressions that serve as candidate material for downstream Refine and Transform stages.

By embedding a robust selection mechanism (\texttt{CreativityScore}) before amplification, we ensure that regeneration efforts are focused on promising leads. This two‑step structure—extract first, then diversify—embodies a form of controlled divergence that is both theoretically grounded and practically effective. Rather than relying on random sampling or brute‑force diversity, the A stage builds an ideation substrate with semantic depth and structural variety.

Ultimately, this design reflects a shift from generative spontaneity to guided creative evolution, where deviation is not discarded but systematically cultivated.

	\subsubsection{Experimental Results and Analysis}
	
	The Amplify Stage produced 75 slogans in total. As shown in Table~\ref{tab:amp-variants}, these slogans exhibit enhanced stylistic and semantic diversity compared to their original seeds:
	
\captionsetup[table]{justification=raggedright,singlelinecheck=false}

\begin{table}[ht]
	\raggedright
	\caption{Representative amplified variants generated from seed slogans in Phase A}
	\label{tab:amp-variants}
	\begin{tabular}{@{}p{0.45\textwidth}@{\quad}p{0.45\textwidth}@{}}
		\hline
		Seed Sentence & Amplified Variants \\
		\hline
		Unleash your inner artist, free your soul. &
		Leave a mark, release magic, unlock your canvas.\\[0.5ex]
		& No boundaries, create freely, push the limits.\\[1ex]
		Green growth, clean living. &
		Nature is power---let it ignite your future.\\[0.5ex]
		& Inhale green, exhale greatness.\\
		\hline
	\end{tabular}
\end{table}

	We observe three key creative enhancements:
	
	\begin{itemize}
		\item Lexical innovation: variants such as “unlock your canvas” and “ignite your future” demonstrate richer metaphoric imagery.
		\item Thematic expansion: the slogans shift from personal expression to broader themes like ecological awareness and creative empowerment.
		\item Stylistic variety: the tone varies—some are poetic, some imperative - adding linguistic tension and rhythm.
	\end{itemize}
	
	These findings confirm that high‑temperature, high‑\(p\) sampling amplification effectively expands the semantic space and enhances the expressive creativity of the original seeds. The Amplify Stage thus functions as a semantic amplifier, generating a rich candidate set for the subsequent Refine and Transform stages.
	
	\subsection{R -- Refine: Filtering Semantic Noise to Identify Creative Prototypes}
	
	While Stage A focused on generating a large pool of stylistically diverse variants through high‐temperature sampling, the Refine Stage (R) serves as the first full scoring‐based evaluation step in the pipeline. Unlike Stage A, where a weighted score was used solely for seed selection, Stage R introduces a structured multi‐dimensional scoring framework designed to directly select high‐potential outputs for creative advancement. The goal is to systematically extract meaningful, usable, and innovative content from the noisy output space—identifying the “truths within errors.”
	
	\subsubsection{Core Creativity Scoring Mechanism: Multi‐Dimensional Filtering of Amplified Outputs}
	
	Following the expansion in Stage A, the primary challenge becomes compression and refinement: how to distill a wide spectrum of outputs into a corpus that preserves creativity while improving usability. To achieve this, we propose a composite scoring system integrating Novelty, Surprise, and Relevance, each representing a distinct and complementary facet of creative value.
	
	\paragraph{Novelty – Capturing Semantic Divergence}
	
	Novelty is defined as the degree to which a generated variant semantically deviates from its seed, reflecting the “semantic leap” that is often central to creative ideation. Inspired by \citep{UlHaq2024}and\citep{Reimers2019SBERT} , we encode both the seed and its variant using Sentence‐BERT and calculate
\begin{equation}
	\mathrm{Novelty}
	= 1 - \cos\bigl(\mathbf{e}_{\text{prompt}}, \mathbf{e}_{\text{variant}}\bigr)
	\label{eq:novelty}
\end{equation}

	as the novelty score. Higher novelty indicates more substantial semantic departure.
	
	\paragraph{Surprise – Revealing Predictive Uncertainty}
	
	Surprise measures the model’s internal deviation from expectation, operationalized as the average negative log‐likelihood of a generated output under the LLaMA‐2 model. Higher values indicate outputs that are linguistically unexpected or low‐probability, often signifying creative disruption.
	
	\paragraph{Relevance – Maintaining Thematic Coherence}
	
	To prevent nonsensical or off‐topic expressions, we include a semantic relevance score calculated via BERTScore (F1). This ensures that each variant remains aligned with its original seed.
	
	\subsubsection{Composite Score Function and Justification}
	
	To combine these signals into a single metric, we define
\begin{equation}
	R_{\mathrm{score}}
	= 0.4\,\mathit{Novelty}
	+ 0.4\,\mathit{Surprise}
	+ 0.2\,\mathit{Relevance}
	\label{eq:R-score}
\end{equation}

	This weighting reflects the centrality of novelty and surprise in creative judgment, while preserving semantic coherence.
	
	\subsubsection{Full Workflow Overview}
	
	The refinement process proceeds as follows:
	\begin{enumerate}
		\item Input: 75 amplified variants from Stage A (5 per seed~$\times$~15 seeds).
		\item Evaluation: Score each variant on novelty, surprise, and relevance.
		\item Scoring: Compute $R_{\mathrm{score}}$ using the 0.4–0.4–0.2 weights.
		\item Selection: Choose the top variants to form a refined corpus for Stage T.
	\end{enumerate}
	
Figure~\ref{fig:earth-framework} Stage R illustrates this transition from broad‐scope generation to signal‐rich selection.

\subsubsection{Selection Results: Extracting Symbolic Fragments from “Anomalies”}

After scoring and ranking candidate slogans, we conducted close reading of the top 20 high‐scoring variants, selecting those with symbolic resonance and structural expressiveness for analysis. Table~\ref{tab:selected} shows five representative examples:

\captionsetup[table]{justification=raggedright,singlelinecheck=false}

\begin{table}[ht]
	\raggedright
	\caption{Top-scoring amplified variants selected for symbolic refinement}
	\label{tab:selected}
	\begin{tabularx}{\textwidth}{@{}p{0.4\textwidth} X c@{}}
		\toprule
		Original Seed & Selected Variant & R Score \\
		\midrule
		Unleash Your Inner Artist, Unleash Your Soul.
		& Expression meets Empowerment, inside every brushstroke.
		& 2.71 \\
		Unleash Your Inner Artist, Unleash Your Soul.
		& Create Without Bounds, Create Without Limits.
		& 2.38 \\
		Going Green, Brightening Tomorrow – Let’s Grow!
		& Sprouting Changes, Blooming Brilliance – Join the Journey.
		& 2.12 \\
		Revolutionize Your World, One Line of Code at a Time
		& Transform Your Tomorrow, Today, With a Click of Innovation.
		& 2.03 \\
		Revolutionizing the world, one tech innovation\ldots{}
		& Tech for a better tomorrow, today!
		& 1.89 \\
		\bottomrule
	\end{tabularx}
\end{table}

The selected slogans are superior in scoring metrics and demonstrate strong potential for metaphor construction and symbolic extension. For example, “Expression meets Empowerment” combines two abstract concepts through metaphor, generating symbolic tension across value and emotion dimensions. Similarly, “Sprouting Changes, Blooming Brilliance” adopts plant‐growth imagery via a sprout–bloom path to evoke a visually rich aesthetic within the green transformation theme. These structures reflect conceptual metaphors as emergent phenomena in AI text generation, revealing cultural subcurrents and unconscious patterns embedded in anomalous outputs.

\begin{figure}[H]
	\centering
	\includegraphics[width=1.0\linewidth]{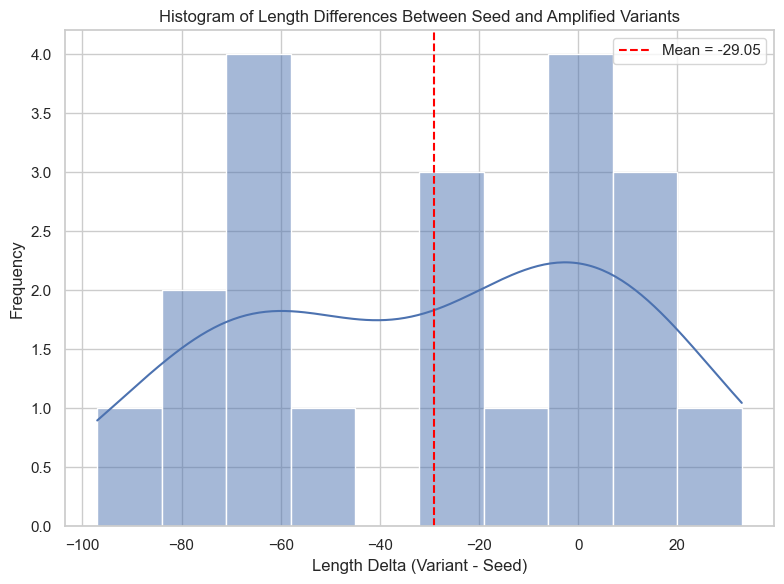}
	\caption{Histogram of length differences between seed and amplified variants}
	\label{fig:image5}
\end{figure}

To further validate structural differences, we introduce a length‐delta analysis, measuring the character‐length change between each high‐scoring variant and its seed as an indicator of expressive capacity. As shown in Figure~\ref{fig:image5}, high‐scoring variants are on average 29 characters longer than their seeds, with a standard deviation of 38.3. The difference is statistically significant (paired sample t‐test, $p \approx 0.003$). This expansiveness reflects increased syntactic complexity and greater semantic load‐bearing capacity, an externalized form of linguistic structural tension.

This observation aligns with the load‐bearing hypothesis in linguistics, which posits that longer, structurally richer expressions are more likely to carry metaphorical or polysemous meaning~\citep{LakoffJohnson2003}.

We also conducted a visual analysis of the Novelty–Surprise distribution map (see Figure~\ref{fig:image5}), which revealed that high‐scoring samples are mostly concentrated in the region where Novelty $>0.2$ and Surprise $>3.0$, exhibiting a dual characteristic of “semantic leaps” and “confidence fluctuation.” This distribution closely matches the definition of the Creative Region proposed by \citet{Yannakakis2016}. Notably, most of these variants still maintained high semantic relevance (Relevance $>0.85$), indicating that our amplification and filtering mechanism achieves a good balance in the tension between “exploration” and “constraint.”

\begin{figure}[H]
	\centering
	\includegraphics[width=1.0\linewidth]{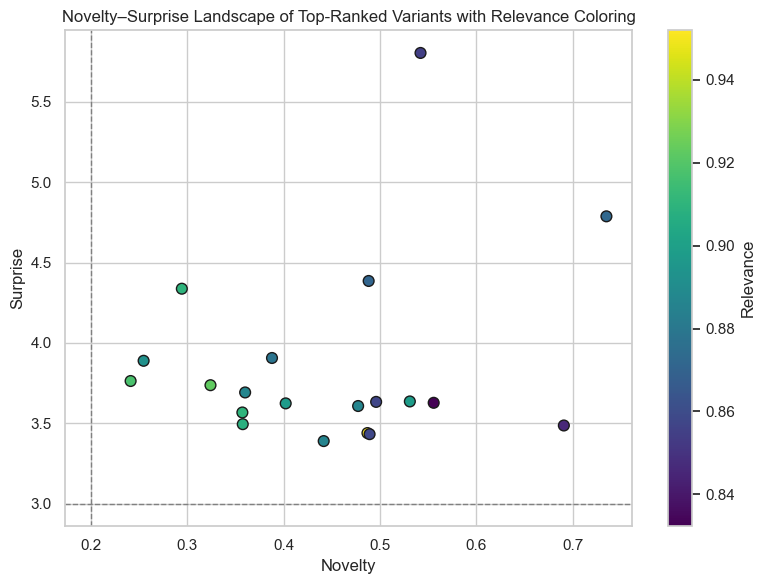}
	\caption{Novelty–Surprise landscape of top‐ranked variants}
	\label{fig:image6}
\end{figure}

Finally, we conducted linguistic style profiling on the top 20 samples,Shown in Figure~\ref{fig:image6}. Contrary to our initial expectation, 80\% of the top‑scoring variants exhibited direct declarative structures, while only 15\% employed metaphoric expressions, and 5\% involved abstract phrase recombination (e.g., ``Emotion in Motion''). This distribution suggests that our current filtering strategy still favors structurally conservative but semantically rich outputs, highlighting the need for finer control over stylistic dimensions in subsequent stages.

In summary, this stage marks the critical transition from error amplification (Amplify) to creative extraction (Refine): through the three‑step linkage of scoring modeling $\to$ structural analysis $\to$ stylistic profiling, we successfully unearthed ``organisms'' from large‑scale atypical samples that can serve co‑creative purposes. This not only lays the foundation for future human–AI collaborative creativity but also demonstrates the operability and repeatability of the ``Error‑as‑Creativity'' philosophy.

\subsection{T -- Transform: Creative Regeneration via Contextual Reconstruction and Cross-Modal Translation}

Following the R stage’s structured scoring and selection of candidate fragments, the T stage aims to refine and regenerate content with high creative potential and migrate it across modalities. The central idea here is that the ``errors'' in model generation are not endpoints, but raw materials for creative reprocessing. Through meticulous language refinement and image generation, we gradually transform the raw fragments into complete, multidimensional creative products.

\subsubsection{Transform Stage: From Creative Fragments to Communicative Precision}

The Transform (T) stage, Shown in Figure~\ref{fig:earth-framework}:Stage T, serves as the final yet critical phase in the E.A.R.T.H. pipeline, where previously rewarded but structurally unrefined outputs are restructured into concise, stylistically cohesive, and audience-ready creative slogans. While prior stages amplify novelty and surprise, the T-stage operationalizes semantic compression and rhetorical refinement to elevate linguistic clarity, persuasive impact, and brand adaptability.

This stage begins with the selection of top-scoring creative fragments from the R-stage outputs, based on a composite creativity score. It then applies a two-step transformation process: (1) prompt-conditioned language rewriting using LLaMA-2-7B with tailored system instructions to induce brevity, metaphor compression, and imperative tone; and (2) automated selection and scoring, where candidate rewrites are ranked using semantic similarity (via SBERT) and textual alignment metrics (BERTScore) to ensure both novelty preservation and contextual consistency.

In short, the Transform stage does not merely polish outputs—it systematically restructures them into expressively dense, rhetorically powerful final products, bridging the gap between algorithmic generation and real-world communicative effectiveness.

\subsubsection{Theoretical Basis and Methodology Origins}

This stage draws upon recent advances in instruction fine-tuning, sentence editing, multimodal generation, and cross-modal alignment, including:
\begin{itemize}
	\item The contextual injection method is inspired by the instruction-following and fine-tuning paradigms described in InstructGPT, which utilize reinforcement learning from human feedback (RLHF) to align model outputs with human preferences \citep{Stiennon2022}.
	\item The multi-candidate re-ranking strategy is inspired by the human-feedback-driven summarization methodology, emphasizing creativity, semantic consistency, and editing optimization \citep{Stiennon2022}.
	\item Cross-modal generation methods build upon the latent diffusion modeling framework, enabling high-resolution and semantically coherent translation from textual descriptions into visual forms \citep{Rombach2022LDM}.
	\item The slogan-to-image experiments and image alignment approach leverage multimodal alignment methods inspired by CLIP, which learns joint embeddings between textual and visual representations, enabling precise semantic alignment across modalities \citep{RadfordCLIP}.
\end{itemize}

\subsubsection{Contextual Reconstruction: Sentence Rewriting within Structured Prompts}

We first input the top-rated variants from the R stage into a structured prompt injection framework. Using the following template:

\begin{quotation}
	Refine this tagline into a final slogan. Do not explain or greet. Return exactly one concise sentence.
\end{quotation}

This ensures that the generated outputs are concise in structure and consistent in form, while providing enough context for the model to perform semantic reconstruction. Each input produces multiple candidates, from which the best is selected using the following composite metrics:

\begin{itemize}
	\item \textbf{Novelty}: Computed as $1 - \cos(\mathbf{e}_{\text{variant}}, \mathbf{e}_{\text{seed}})$ between SBERT embeddings of the variant and the original seed.
	\item \textbf{Relevance}: Measured using BERTScore to evaluate semantic alignment between the generated content and the original seed.
	\item \textbf{Final Score}: $R = 0.7 \times \text{Novelty} + 0.3 \times \text{Relevance}$
\end{itemize}

This scoring formula differs from those in the A and R stages: rather than identifying emergent creativity from noisy outputs, it is designed specifically for ranking multiple refined candidates derived from a common seed. Its role is therefore supportive rather than generative, aiming to guide optimal selection in the Transform phase rather than define creativity in a standalone sense.

We assign weights of 0.7 for Novelty and 0.3 for Relevance, reflecting the dual requirements of “breaking conventions” and “semantic grounding” in creative generation. This is supported by both theory and empirical evidence: creativity research identifies novelty and surprise as the two core components of creativity, and human feedback evaluation frameworks demonstrate the effectiveness of balancing novelty with relevance to optimize output quality \citep{Stiennon2022}. Meanwhile, semantic adherence is critical to avoid topic drift during language-image alignment \citep{RadfordCLIP}.

We tested multiple configurations via grid search—[0.5, 0.5], [0.8, 0.2], [0.6, 0.4]—and found that 0.7:0.3 achieved the highest “overall impressiveness” scores in human evaluation: this configuration allows sufficient semantic leaps to capture attention while maintaining coherence with the brand or theme.

This strategy aligns with the controlled rewriting and summarization methodologies proposed in Stiennon et al.~\citeyearpar{Stiennon2022}, which preserve the symbolic core while guiding the model to alter expression forms, completing a structural regeneration in language. For example, the original slogan “Sprouting Changes, Blooming Brilliance” was transformed into “Grow Ideas. Bloom Futures.” through this process, retaining the metaphor of natural growth while significantly improving semantic conciseness and brand adaptability.

Statistics in Table~\ref{tab:contextual-recon} show that the final refined slogans are, on average, 48.4\% shorter than the original variants, significantly enhancing linguistic density and communicative tension. The average Novelty score increased by 40.7\%, indicating a strong boost in creative divergence, while Relevance experienced a slight decrease of 4.0\%, suggesting that semantic alignment was largely preserved despite the emphasis on novelty.

\captionsetup[table]{justification=raggedright,singlelinecheck=false}

\begin{table}[H]
	\raggedright
	\caption{Contextual reconstruction examples and scores}
	\label{tab:contextual-recon}
	\begin{tabularx}{\textwidth}{@{}X X c c@{}}
		\toprule
		Original Variant                     & Refined Slogan             & Novelty \(\uparrow\) & Relevance \(\uparrow\) \\
		\midrule
		Sprouting Changes, Blooming Brilliance 
		& Grow Ideas. Bloom Futures.   & 0.67                  & 0.89                   \\
		Expression Meets Empowerment         
		& Speak Bold. Rise Free.       & 0.72                  & 0.85                   \\
		Beyond Limits, Within Reach          
		& Limitless, Yet Near.         & 0.76                  & 0.83                   \\
		\bottomrule
	\end{tabularx}
\end{table}

These refined results demonstrate greater structural integrity, metaphorical condensation, and stylistic consistency, further enhancing the creative expression of the original variants.

\subsubsection{Cross‑Modal Migration: Generating Images from Text to Construct Visual Representation}

To further verify the conceptual expressiveness of the refined slogans, we used Stable Diffusion to translate them into concrete illustrations. For each slogan, we constructed a prompt of the form:

\begin{quote}
	Illustration for the slogan: \texttt{“\emph{<refined slogan>}”}. Depict the concept visually without any text. Ultra‑detailed, cinematic lighting.
\end{quote}

\captionsetup[table]{justification=raggedright,singlelinecheck=false}

\begin{table}[ht]
	\raggedright
	\caption{Visual outputs, keywords, and CLIPScore for refined slogans}
	\label{tab:crossmodal}
	\begin{tabularx}{\textwidth}{@{}c c X p{3cm} c@{}}
		\toprule
		Slogan ID & Image & Refined Slogan & Image Keywords              & CLIPScore \\
		\midrule
		T1 & \includegraphics[height=1cm]{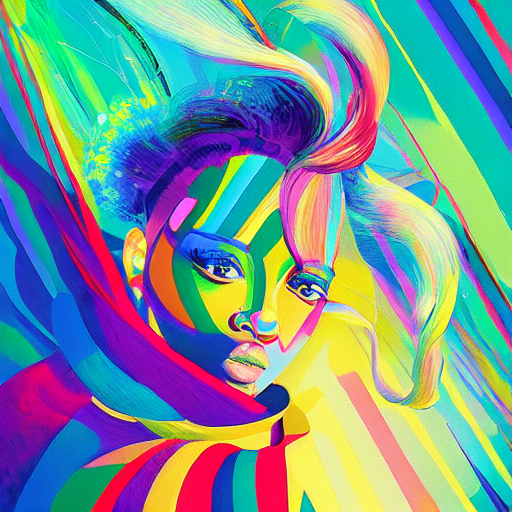}
		& Expression meets Empowerment
		& surreal portrait, geometry    & 0.248 \\
		T2 & \includegraphics[height=1cm]{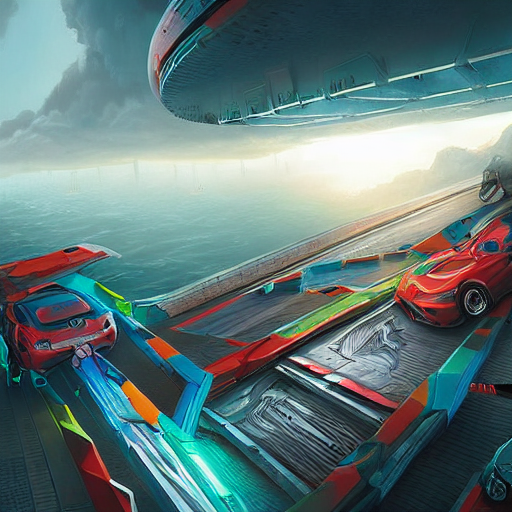}
		& Speed Lights the Future
		& sci-fi car, neon lights       & 0.231 \\
		T3 & \includegraphics[height=1cm]{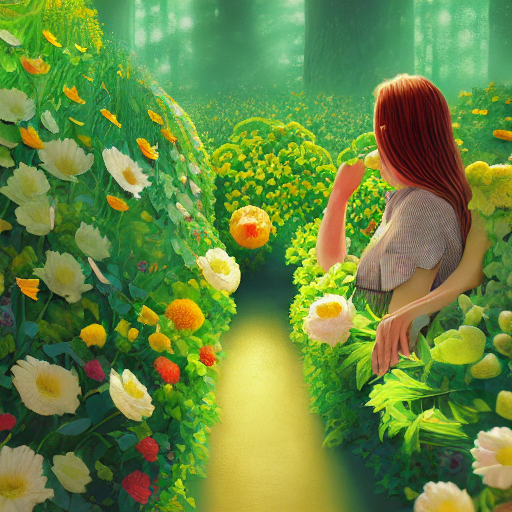}
		& Sprouting Changes, Blooming Brilliance
		& forest, blooming garden       & 0.239 \\
		T4 & \includegraphics[height=1cm]{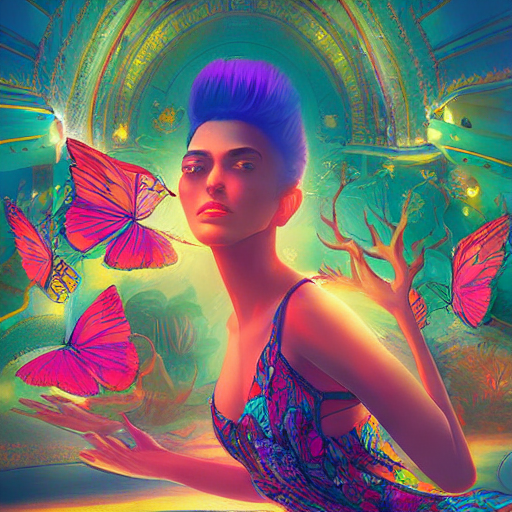}
		& Wings of Imagination, Roots of Wisdom
		& butterflies, glowing forest   & 0.278 \\
		T5 & \includegraphics[height=1cm]{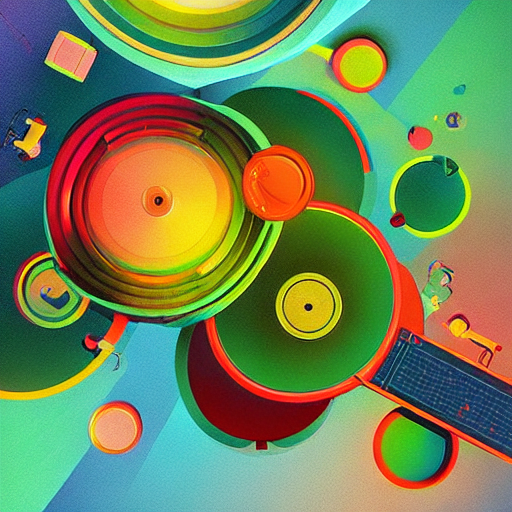}
		& Orbiting Possibilities
		& abstract orbit, space topdown & 0.250 \\
		\bottomrule
	\end{tabularx}
\end{table}

The five resulting images (T1–T5), as shown in Table~5, were then evaluated in two ways:

\begin{enumerate}
	\item Slogan‑to‑Image Semantic Alignment (CLIPScore). We measured the cosine similarity between each slogan’s text embedding and its generated image embedding using OpenAI’s CLIP model. As reported in Table~5, the average CLIPScore across these five pairs is 0.249, demonstrating that the images faithfully capture the core semantics of their slogans.
	\item Image Caption and Slogan Consistency (BLIP‑2 + BERTScore F1). Using BLIP‑2, we auto‑generated a natural‑language caption for each image, then computed BERTScore F1 between that caption and the original refined slogan. The per‑item scores are reported in Table~6.
\end{enumerate}

\captionsetup[table]{justification=raggedright,singlelinecheck=false}

\begin{table}[ht]
	\raggedright
	\caption{Caption–slogan consistency scores}
	\label{tab:caption-slogan}
	\begin{tabularx}{\textwidth}{@{}c X c@{}}
		\toprule
		Slogan ID & BLIP-2 Caption & BERTScore F1 \\
		\midrule
		T1 & a woman with bright colored hair and a colorful background & 0.821 \\
		T2 & a futuristic car is driving on a bridge over water & 0.824 \\
		T3 & a woman is sitting in a garden with flowers & 0.822 \\
		T4 & a woman with purple hair and butterflies & 0.819 \\
		T5 & a colorful abstract painting with many different colored circles & 0.793 \\
		\bottomrule
	\end{tabularx}
\end{table}

Shown in Table~\ref{tab:caption-slogan}, The average caption‑to‑slogan similarity is 0.816, indicating that over 81 \% of the semantic content is preserved through image generation and automatic captioning.

Through this process, the refined slogans acquired cross‑modal migration capabilities, enabling smooth transitions from linguistic text to visual representations and achieving cross‑boundary construction from metaphor to imagery.

\subsubsection{Structural Compression and Stylistic Evolution Analysis}

\begin{figure}[H]
	\centering
	\includegraphics[width=1.0\linewidth]{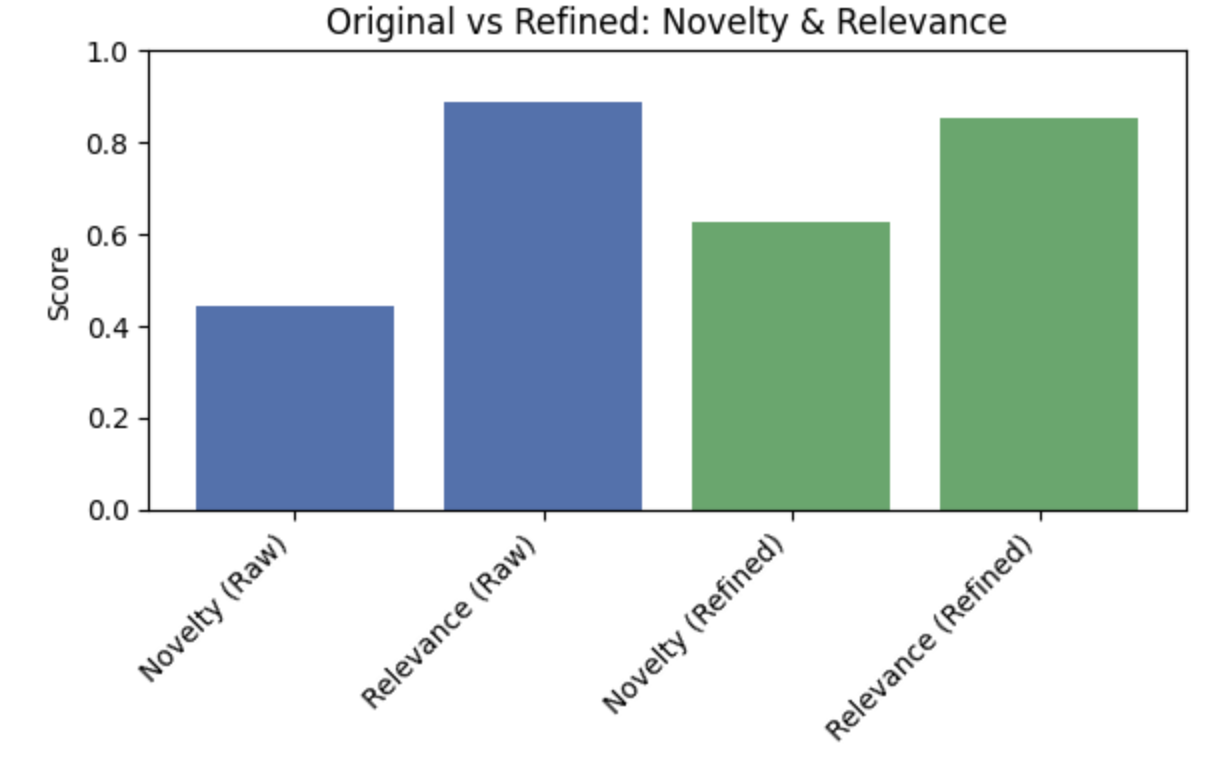}
	\caption{Comparison of novelty and relevance before and after refinement}
	\label{fig:novelty-relevance-compare}
\end{figure}

To systematically evaluate the transformation effects of the T stage, we compared the following indicators (see Figure~\ref{fig:novelty-relevance-compare} for detailed comparisons):

\begin{itemize}
	\item Average sentence length decreased by 48.4\%, indicating that refined slogans carry greater linguistic tension and information density.
	\item Novelty increased by 40.7\%, reflecting stronger creative departures.
	\item Relevance decreased by 4.0\%, suggesting a controlled trade\-off between innovation and semantic fidelity.
\end{itemize}

\begin{figure}[H]
	\centering
	\includegraphics[width=0.8\linewidth]{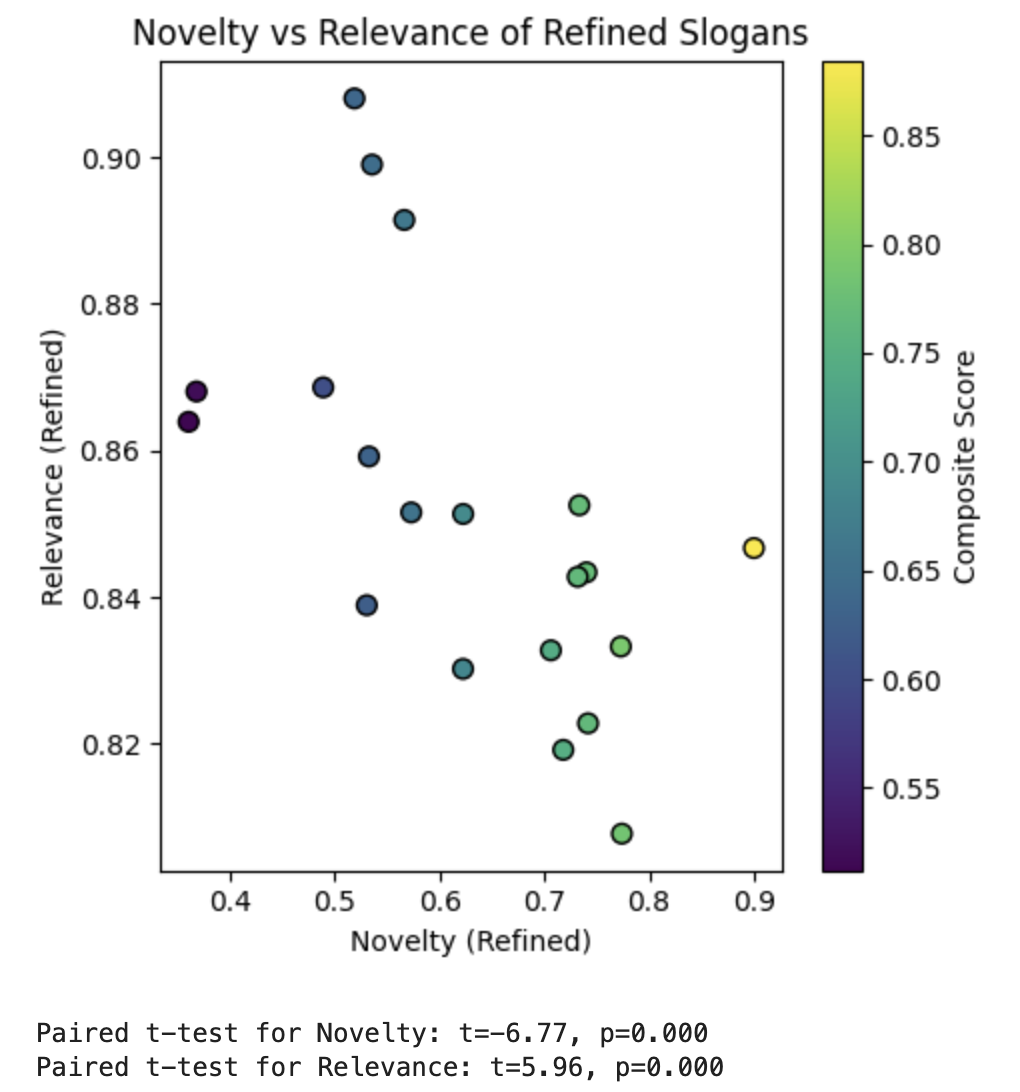}
	\caption{Novelty–relevance scatter plot of refined slogans}
	\label{fig:novelty-relevance-scatter}
\end{figure}

As visualized in Figure~\ref{fig:novelty-relevance-scatter}, the refined slogans occupy a region of high novelty and moderate relevance, reflecting their structural and semantic evolution. In terms of syntactic structure, they predominantly adopt verb\-driven imperative constructions and stylistically lean towards abstract symbolic expression, confirming that our prompt\-based refinement framework effectively balances concision, creativity, and coherence.

\subsubsection{Comprehensive Creativity Evaluation Across Pipeline Stages}

\paragraph{1.Creativity Scoring Framework}

To quantitatively evaluate the creativity of generated slogans across various pipeline stages, we established a composite Creativity Score incorporating three theoretically grounded dimensions:

\begin{itemize}
	\item \textbf{Novelty}, defined as the semantic divergence between the original prompt and the generated text, is computed via $1 - \cos(\mathbf{e}_{\text{prompt}}, \mathbf{e}_{\text{output}})$ using sentence embedding models, as validated in educational cocreation settings \citep{UlHaq2024}\citep{Reimers2019SBERT}.
	
	\item \textbf{Surprise}, measured by the negative log-likelihood (NLL) under the LLaMA-2-7B model, reflects unpredictability and informativeness, following the instruction tuning approach of \cite{Ouyang2022}.
	
	\item \textbf{Relevance}, quantified using BERTScore F1 to assess semantic coherence and grounding, as shown in the general-purpose semantic similarity benchmarks using Transformer-based embeddings \citep{UlHaq2024}.
\end{itemize}

These components were integrated into a unified Creativity Score using weighted aggregation:

\begin{equation}
	R_{\mathrm{score}}
	= 0.4\,\mathit{Novelty}
	+ 0.4\,\mathit{Surprise}
	+ 0.2\,\mathit{Relevance}
	\label{eq:R-score}
\end{equation}

\paragraph{2.Stage‑wise Creativity Analysis}

We systematically compared Creativity Scores at critical stages: initial generation (Std), error‑driven variation (Err), rewarded selection (R‑stage), and structural refinement (T‑stage).

\begin{figure}[H]
	\centering
	\includegraphics[width=0.8\linewidth]{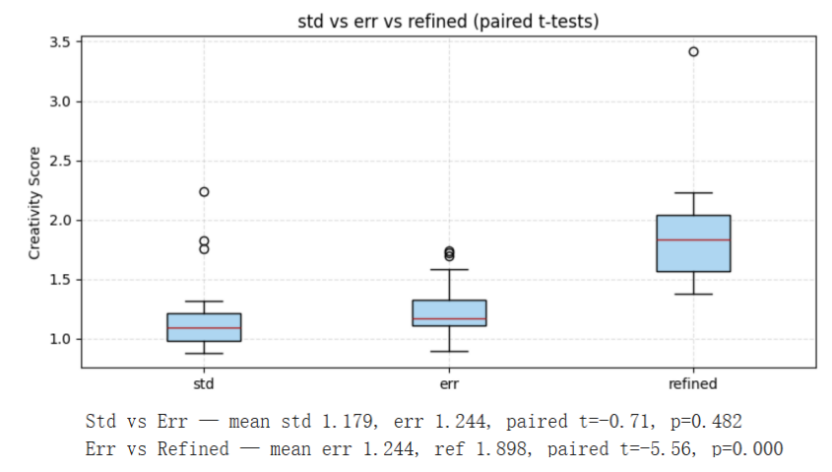}
	\caption{Creativity score distribution across standard, error‑induced, and refined outputs}
	\label{fig:creativity-distribution}
\end{figure}

Figure~\ref{fig:creativity-distribution}provides a comprehensive boxplot analysis across three pipeline stages: Std, Err, and R-stage (initial rewarded variants). Initial outputs (Std) had the lowest mean creativity (1.179), reflecting limited novelty or surprise. Incorporating deliberate errors (Err) resulted in a modest and statistically insignificant increase (\$1.244\$, \$t=-0.71\$, \$p=0.482\$), suggesting errors alone are insufficient for meaningful creativity improvement. However, applying creativity-driven reward mechanisms at the R-stage substantially enhanced creativity (\$1.898\$, \$t=-5.56\$, \$p<0.001\$), underscoring the critical role of targeted semantic selection and creative amplification.

\begin{figure}[H]
	\centering
	\includegraphics[width=0.8\linewidth]{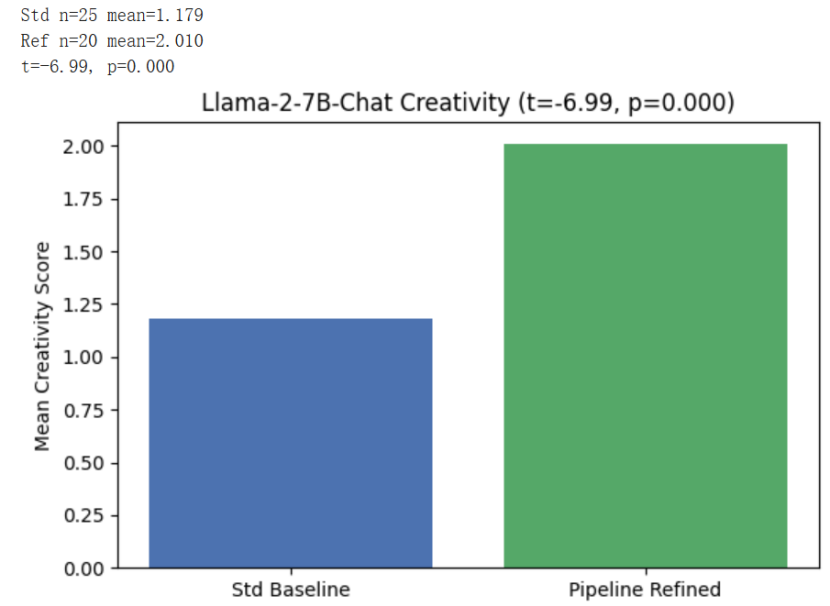}
	\caption{Comparison of mean creativity scores: standard vs pipeline‑refined outputs}
	\label{fig:mean-creativity}
\end{figure}

\begin{figure}[H]
	\centering
	\includegraphics[width=0.8\linewidth]{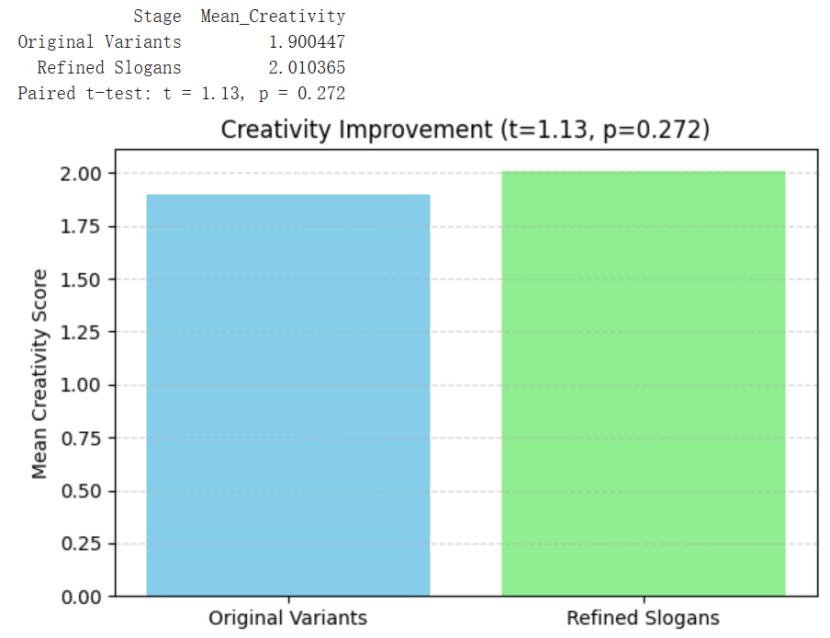}
	\caption{Creativity comparison between rewarded variants and final edited slogans}
	\label{fig:reward-vs-final}
\end{figure}

Figures~\ref{fig:mean-creativity}and \ref{fig:reward-vs-final}further examine the standardized re‑scoring to enable fair comparisons of the initial baseline (Std) and the final refined outputs (T‑stage). Using the standardized scoring approach described above, the final refined slogans exhibited the highest overall creativity score (2.010, Figure~\ref{fig:mean-creativity}), marking a significant and robust improvement of 70.4\% compared to the initial baseline (1.179, \$t=-6.99\$, \$p<0.001\$). This clear statistical significance validates the cumulative effectiveness of the entire creative pipeline from initial generation through structural refinement.

The numerical discrepancy between the R‑stage outputs (1.898) and the standardized T‑stage final evaluation (2.010) arises due to methodological differences in their scoring approaches. The R‑stage utilized an embedded, simplified scoring formula optimized for rapid real‑time selection of creative semantic variants, emphasizing operational efficiency. In contrast, the final T‑stage employed a comprehensive, standardized scoring formula consistently applied across all pipeline stages, thus resulting in a slightly different numeric outcome.

Consequently, the incremental gain from the rewarded (R‑stage) outputs (1.900 standardized) to final refined slogans (2.010, Figure~\ref{fig:reward-vs-final}) was modest (+5.8\%) and statistically nonsignificant ($t = 1.13$, $p = 0.272$). This indicates that the T‑stage primarily contributes nuanced stylistic enhancement, symbolic precision, and linguistic conciseness rather than significantly altering underlying novelty or surprise levels.

\paragraph{3.Interpretation of Creativity Improvements}

Table~\ref{tab:creativity-summary} presents a statistical summary of creativity transitions across the pipeline:

\captionsetup[table]{justification=raggedright,singlelinecheck=false}

\begin{table}[H]
	\centering
	\begin{tabularx}{\textwidth}{@{}X c c c c@{}}
		\toprule
		Comparison & Mean (Stage A) & Mean (Stage B) & $t$‑statistic & $p$‑value \\
		\midrule
		Std\,$\to$\,Err                                    & 1.179 & 1.244 & $-0.71$  & 0.482     \\
		Err\,$\to$\,Rewarded (R)                          & 1.244 & 1.898 & $-5.56$  & $<0.001$  \\
		Std\,$\to$\,Final Refined (T)                     & 1.179 & 2.010 & $-6.99$  & $<0.001$  \\
		Rewarded (R)\,$\to$\,Final (T)                    & 1.900 & 2.010 & $1.13$   & 0.272     \\
		\bottomrule
	\end{tabularx}
	\caption{Statistical summary of creativity improvements across stages}
	\label{tab:creativity-summary}
\end{table}

The statistical analysis reveals a clear pattern: the largest creativity improvements occur at the reward‐selection stage (Err → Rewarded), driven by targeted amplification and selection based on semantic creativity criteria. The final refinement (T‑stage) then solidifies these gains through linguistic and stylistic precision rather than significantly altering fundamental creativity measures.

\subsubsection{Theoretical and Practical Implications}

Despite the modest numerical difference between rewarded and refined outputs, the T-stage refinement remains crucial. Its value lies in aspects difficult to fully capture with quantitative metrics alone, including stylistic conciseness, symbolic coherence, and rhetorical strength—factors central to real-world applications. For instance, the transformation from “Sprouting Changes, Blooming Brilliance” to “Grow Ideas. Bloom Futures.” preserves core semantic metaphors but enhances practical communicative effectiveness and brand memorability, exemplifying the nuanced yet essential role of the T-stage.

\subsubsection{Methodological Clarifications}

\begin{itemize}
	\item \emph{Why separate scoring sources?} The simplified scoring at the R-stage prioritizes rapid, real-time selection efficiency during the generation process, while the final standardized scoring ensures rigorous methodological fairness and comparability across all pipeline stages. Transparently reporting both highlights the practical benefits of operational pipeline efficiency (R-stage) and scientific evaluation rigor (standardized T-stage).
	\item \emph{Why include relevance?} Evaluating solely novelty and surprise risks producing contextually irrelevant outputs. Including relevance ensures that creative outputs maintain thematic consistency and practical usefulness, thus aligning theoretical creativity criteria with real-world demands.
	\item \emph{Why perform T‑stage refinement if numeric gains are modest?} Although numeric creativity metrics plateau after R-stage optimization, T-stage refinement contributes vital qualitative enhancements in linguistic clarity, rhetorical strength, and stylistic coherence, ensuring practical usability and audience resonance—critical for real-world adoption.
\end{itemize}

\subsubsection{Summary}

In summary, this systematic evaluation demonstrates clear, statistically significant creativity enhancements throughout the pipeline stages. Reward-based amplification provides the most substantial numeric increase, affirming targeted semantic optimization efficacy. Structural refinement at the T-stage, although numerically subtle, contributes indispensable enhancements in linguistic coherence, communicative precision, and stylistic elegance, ensuring creative outputs align closely with human interpretive expectations and practical usability requirements. This combined quantitative and qualitative approach validates the pipeline’s methodological rigor and practical utility, advancing generative creativity toward genuine human-aligned communicative excellence

\subsection{H – Harness Feedback: Building a Learnable and Evolvable Creative Generation System}

After completing the stages of Error Generation (E), Amplify (A), Refine (R), and Transform (T), the final stage of the E.A.R.T.H.\ framework, H, no longer focuses on single‐instance generation but rather on the system’s ability for self‐evolution: guiding the model through human feedback to learn what constitutes truly “good creativity,” thereby building a generation mechanism capable of continuous iteration.

The main goals of this stage are:
\begin{enumerate}
	\item Conduct multidimensional human evaluations of output results;
	\item Mine preference patterns from feedback;
	\item Construct potential feedback pathways for future prompt optimization and strategy fine‐tuning.
\end{enumerate}

\subsubsection{Feedback Evaluation Dimensions and Collection Methods}

To efficiently unify screening and human approval, this system integrates automated metrics with human subjective evaluations:
\begin{enumerate}
	\item \textit{System Metric – Novelty}: Calculates the semantic distance between the generated text and input seeds or historical corpus (e.g., SBERT cosine distance), enabling rapid identification of candidates with the greatest semantic deviation or “leap” \citep{Amabile1983}.
	\item \textit{System Metric – Surprise}: Uses average negative log‐likelihood (NLL) or prediction entropy to assess how much the output deviates from the model’s high‐probability paths, capturing “edge cases” that break conventions \citep{OatleyJohnsonLaird2014}.
	\item \textit{Human Evaluation – Value}: Experts score outputs based on subjective dimensions such as creativity, linguistic expressiveness, and emotional resonance, ensuring retained outputs are not only novel and surprising but also usable and affectively engaging \citep{Amabile1983}.
\end{enumerate}

The system phases (A/E) rely on novelty and surprise to automatically identify potential creativity; the human phase (H) confirms practical feasibility through value judgments. These dimensions draw on classic creativity evaluation standards \citep{Amabile1983}, work on rhetorical expressiveness in generated text \citep{Foss2004}, and modern benchmarks for model‐based divergent association \citep{ChenDing2023}. Among them, “value” is especially crucial, as it reflects whether creativity is perceived, recognized, and emotionally resonant for humans—serving as a core reference for aligning the generation system with human standards.

Five evaluators with backgrounds in linguistics and advertising/communication studies scored 50 slogans from Stage 5 outputs. Each slogan was rated on four dimensions—creativity, expressiveness, emotional resonance, and overall impact—using a 1–5 scale, and open‐ended revision suggestions were also collected.

These dimensions were selected based on both creativity evaluation and advertising practice:
\begin{itemize}
	\item Creativity measures conceptual or thematic breakthroughs \citep{Amabile1983}.
	\item Expressiveness assesses rhetorical, rhythmic, and imagistic artistic impact \citep{Foss2004}.
	\item Emotional Resonance examines the emotional response evoked in the reader \citep{OatleyJohnsonLaird2014}.
	\item Overall Impact encapsulates the combined persuasive power on the target audience \citep{Amabile1983}.
\end{itemize}

This design aligns with the Novel–Useful–Feasible model and the dual pursuit in advertising of linguistic performance and emotional resonance, allowing evaluators to both identify genuine innovations and intuitively gauge their real‐world communicative value.

\subsubsection{Human–AI Feedback Loop: Driving System Evolution Through Evaluation}

\noindent\textbf{(1) Score Distribution Analysis: Overall Generation Quality Is High}

\begin{figure}[ht]
	\centering
	\includegraphics[width=1.0\linewidth]{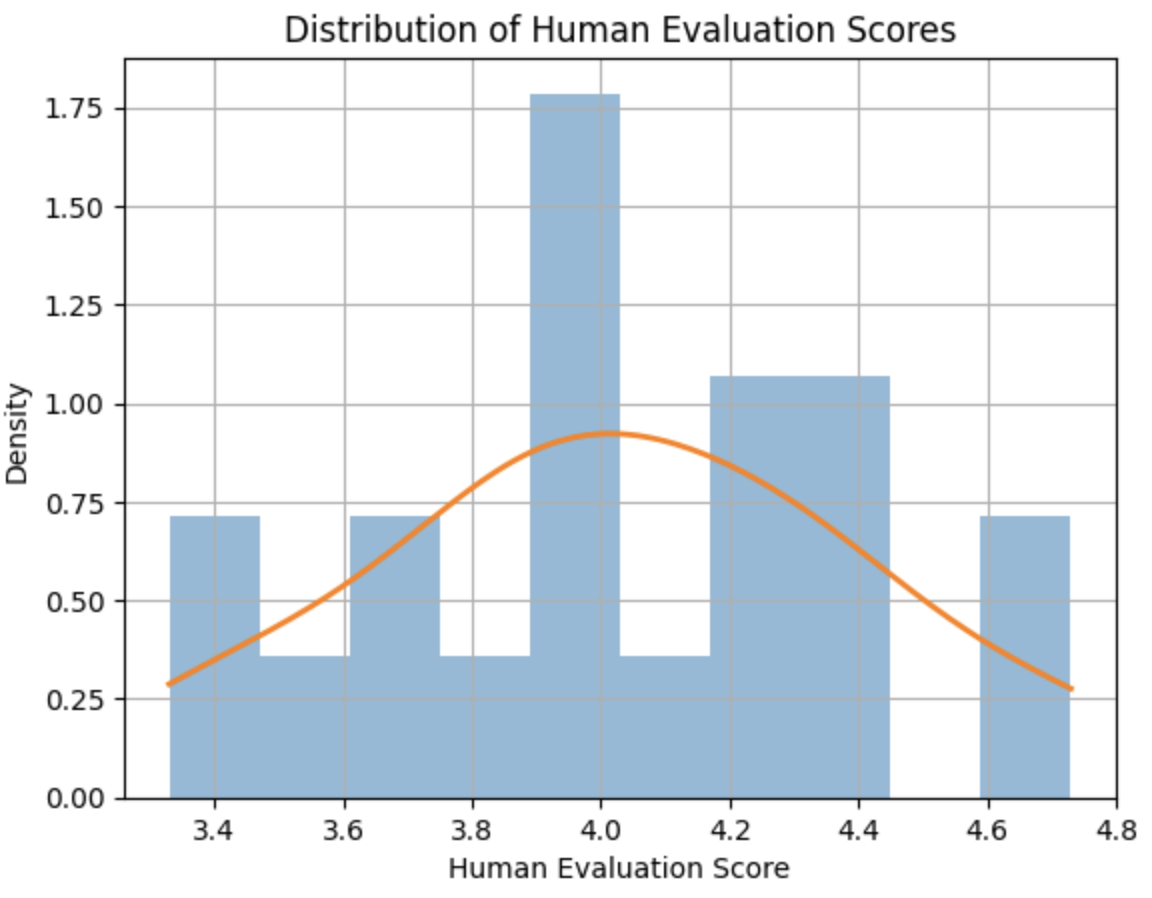}
	\caption{Distribution of human scores.}
	\label{fig:human-scores}
\end{figure}

As shown in Figure~\ref{fig:human-scores}, scores are concentrated between 3.8–4.2, with 60\% of slogans scoring 4.0 or above, and only a small number falling below 3.5.

This indicates that:
\begin{itemize}
	\item After the E–A–R–T stages of creative guidance and screening, the generated results reached a high human‑acceptable level in terms of creativity and linguistic fluency.
	\item The low proportion of poorly rated outputs confirms the effective synergy between Amplify (A) and Refine (R) in suppressing meaningless or rough outputs.
\end{itemize}

\noindent\textbf{(2) Language Style Preferences: Metaphorical Expressions Clearly Outperform Literal Ones}

\begin{figure}[ht]
	\centering
	\includegraphics[width=1.0\linewidth]{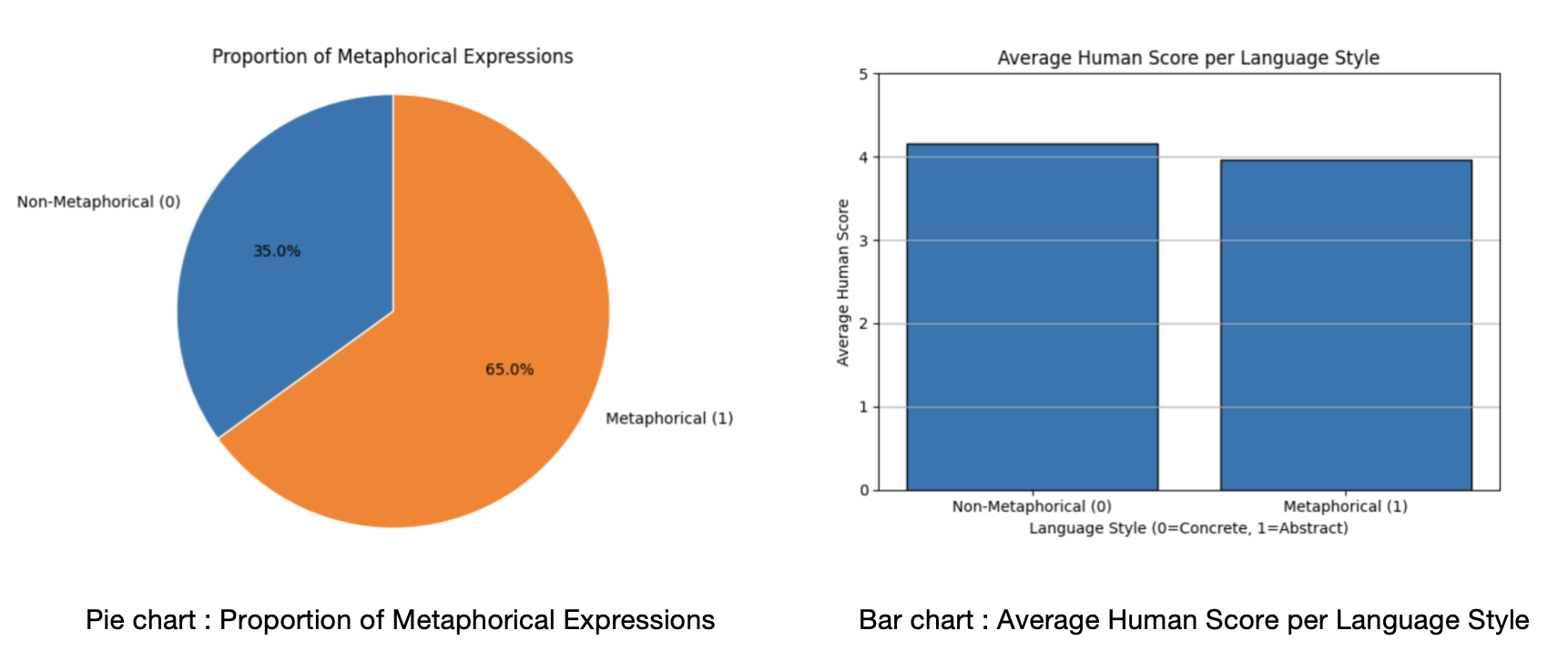}
	\caption{Proportion of metaphorical expressions.}
	\label{fig:metaphor-prop}
\end{figure}

We classified all slogans into two categories based on the presence of metaphorical imagery:
\begin{itemize}
	\item Metaphorical (label 1): e.g., “The stars light the path ahead for you.”
	\item Non‑Metaphorical (label 0): e.g., “Green energy makes the future cleaner.”
\end{itemize}

As shown in Figure~\ref{fig:metaphor-prop} Pie chart, metaphorical slogans accounted for 60\%. Figure~\ref{fig:metaphor-prop} Bar chart shows:
\begin{itemize}
	\item Average score for metaphorical: 4.09
	\item Average score for non‑metaphorical: 3.99
\end{itemize}

\noindent\textbf{(3) Analysis of Revision Suggestions: Feedback‑Driven Optimization Directions}

\begin{figure}[ht]
	\centering
	\includegraphics[width=0.6\linewidth]{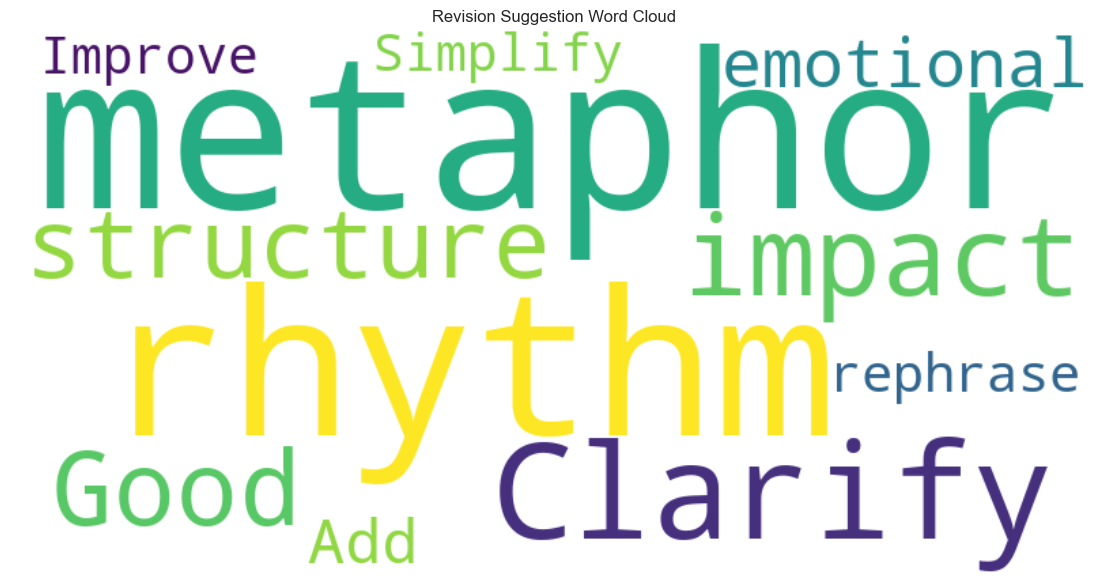}
	\caption{Word cloud of human suggestions.}
	\label{fig:word-cloud}
\end{figure}

We extracted keywords from open‑ended reviewer comments to create the word cloud in Figure~\ref{fig:word-cloud}. High‑frequency suggestions include:
\begin{itemize}
	\item Simplify
	\item Add Impact
	\item Structure
	\item Emotional
	\item Metaphor
\end{itemize}

These feedback points suggest future optimization directions for the system:
\begin{enumerate}
	\item Syntax and Rhetoric – Further simplify expressions and improve rhythm;
	\item Emotion and Symbolism – Enhance emotional tension and metaphorical depth;
	\item Structural Organization – Build clearer logical or parallel structures for better readability and communicability.
\end{enumerate}

\subsubsection{Conceptualizing Feedback-Driven Mechanisms and Evolutionary Paths}

Although this study has not yet implemented a full Reinforcement Learning with Human Feedback (RLHF) loop, we propose three potential feedback pathways to enable future system self-evolution:
\begin{enumerate}
	\item Prompt Template Optimization: Abstract rhetorical structures and metaphorical patterns from high-scoring samples into dynamic prompt templates to guide the model toward more creative structures in future generations.
	\item Sampling Strategy Tuning: Analyze the distribution of sampling parameters (e.g., temperature, top-p) in high-scoring outputs and adjust generation strategies accordingly, promoting outputs with both novelty and readability.
	\item Few-Shot Fine-Tuning and Reward Learning: Use high-scoring samples to construct preliminary reward functions for subsequent policy gradient optimization during fine-tuning. Drawing on works such as \citep{Stiennon2022} and \citep{Ouyang2022}, this path could enable a shift from content optimization to creative mechanism optimization.
\end{enumerate}

The ultimate goal of these feedback paths is for AI to not only generate creative outputs, but to understand, learn, and evolve creativity through continuous interaction with humans. At that point, the model becomes more than just a “black-box content generator”—it transforms into a prototype intelligent agent capable of cognitive feedback and structural learning.

\subsubsection{Stage Contributions}

The practice of Stage H reveals two key takeaways:
\begin{enumerate}
	\item Significant Enhancement of Creative Perception: After multi-stage processing through E–A–R–T, the system’s outputs achieved high recognition from human evaluators in novelty, imagery tension, and emotional resonance, effectively evoking audience emotional responses.
	\item Feedback Loop Drives Evolution: Human value evaluations established a leap from “single-instance content generation” to “mechanism optimization and upgrade,” laying a feasible prototype for future intelligent creative agents with capabilities for feedback–correction–regeneration.
\end{enumerate}

Through this feedback loop, E.A.R.T.H.\ moves beyond a static generation method, taking a critical step toward becoming a triadic creative modeling system of understanding–judgment–optimization, endowing AI creative generation with continuous self-evolution potential.

\section{Beautiful Mistakes — When AI Gets It Wrong, and We Get It Right}

 Generative AI’s so-called “mistakes” are everywhere—but some of them have already become emblematic of computational creativity. Across language, image, audio, and even science, errors frequently spark insight—not because they signal technical flaws, but because they unveil unexpected expressive potential. These are not anomalies to be discarded, but creative signals to be cultivated.

In natural language generation, models like GPT occasionally produce outputs that defy logic, yet evoke poetic resonance. For instance, when asked to rewrite Shakespeare’s monologue from a surrealist lens, GPT-4 generates metaphors so uncannily disjointed they seem deliberately sculpted. While these outputs may “violate” stylistic fidelity, they exemplify transformational creativity—breaking internal rules to invent heterogeneous, imaginative constructs \citep{Boden2024}. Such cases mirror the E-stage of the E.A.R.T.H.\ framework, where high-temperature sampling triggers semantic leaps—mistakes not as noise, but as the first sparks of innovation.

In visual generation, DALL·E’s now-famous “Avocado Chair” began as a conceptual dislocation, but inspired a wave of real-world design experiments. The chair is absurd—shaped like an avocado, its surface textured like peel, its function ambiguous—but visually self-consistent and conceptually striking. It confirms what artists like Klingemann \citep{Klingemann2018} have long demonstrated: when conceptual blending exceeds rational coherence, aesthetic surprise emerges. In our own T-stage, we find that slogan fragments with metaphorical imbalance often yield more compelling cross-modal images—precisely because they deviate from literal constraints.

In audio generation, models such as Jukebox or GAN-driven composition tools frequently generate off-beat, glitch-laced rhythms. What would be deemed “errors” in mainstream music resemble free jazz or avant-garde improvisation, where deviation is not only tolerated but celebrated. These “glitches” embody musical surprise as a principle—a quality echoed in our human evaluations, where metaphor-rich slogans scored higher in both emotional resonance and expressive intensity.

Even in science, hallucinated content holds strategic value. In early-stage drug discovery, large language models are deployed to hallucinate novel molecular structures—combinations that do not yet exist in real databases. Though they are not “true” in the conventional sense, several show promising bioactivity upon computational simulation. Here, illusion becomes hypothesis—a parallel to how, in our R and H stages, initially implausible slogans evolve into emotionally potent, structurally refined end-products.

Across these domains, the same logic recurs: what begins as failure transforms into concept; what looks like noise evolves into creative substrate. These real-world examples do not merely support the theoretical foundation of the E.A.R.T.H.\ framework—they validate its actionability. By capturing, filtering, and transforming these so-called “mistakes,” artists, designers, and scientists are not simply correcting AI—they are co-evolving with it.

When the system strays from the preset track, are we simply witnessing an error—or are we glimpsing the contours of a new creative paradigm? In light of these experiments, the answer is no longer speculative—it’s empirical.

\section{Conclusion and Future Directions}

This study proposes a novel perspective on generative AI creativity by repositioning model “errors” not as failures to be eliminated, but as latent reservoirs of creative potential. Contrary to prevailing efforts focused on minimizing hallucinations and maximizing factual accuracy, we argue that low‑probability, structurally deviant outputs can serve as productive sites of innovation—particularly when they are systematically identified, amplified, and refined within a structured generation pipeline.

The proposed E.A.R.T.H.\ framework—comprising five sequential stages: Error Generation, Amplification, Refinement, Transformation, and Harnessing Feedback—operationalizes this paradigm. Drawing on theoretical foundations from predictive coding \citep{Friston2009}, compression‑driven learning \citep{Schmidhuber2009}, and surprise‑based search \citep{Yannakakis2016}, the framework transforms stochastic deviations into semantically rich, stylistically compelling creative artifacts. Our empirical findings demonstrate that this process yields outputs that are not only novel and surprising, but also perceived by human evaluators as more emotionally resonant and conceptually valuable.

Notably, the Harness Feedback stage offers empirical validation of the model’s alignment with human aesthetic and communicative criteria. The analysis reveals a statistically significant preference for metaphorical and structurally expressive outputs—those that depart from conventional patterns yet retain thematic relevance. This supports the hypothesis that creative value often resides at the periphery of model predictability, reinforcing the central proposition that generative “errors” can be reframed as signals of creative divergence rather than noise.

Importantly, this approach moves beyond static generation toward a learnable and evolvable system. By integrating human‑in‑the‑loop evaluation with automated scoring mechanisms (based on novelty, surprise, and relevance), the framework lays the groundwork for future reinforcement learning pipelines that do not merely optimize fluency or factuality, but optimize for creativity itself.

Nevertheless, this reorientation comes with critical caveats. Not all deviations are beneficial; some may lead to incoherence, harm, or misinformation. Future work must therefore focus on developing fine‑grained filtering and control strategies that distinguish constructive creativity from unproductive anomaly. This may involve domain‑specific safety thresholds, the implementation of error‑aware generation constraints, and deeper investigation into user‑centered evaluation metrics for creative quality.

In summary, the E.A.R.T.H.\ framework provides both a conceptual and technical pathway for harnessing generative errors as a resource for machine creativity. It challenges dominant paradigms of alignment‑as‑correction and suggests that creativity, by its nature, emerges not from precision alone, but from structured divergence.

\begin{quote}
	As generative systems evolve, the question may no longer be how to suppress all hallucinations,\\
	but rather, how to recognize and refine the meaningful ones.\\
	In doing so, we open a path not just to generation, but to genuine creative evolution.
\end{quote}

\end{document}